# Water Quality Data Imputation via A Fast Latent Factorization of Tensors with PID-based Optimizer

Qian Liu, Lan Wang, Bing Yang* and Hao Wu

*Abstract*—Water quality data can supply a substantial decision support for water resources utilization and pollution prevention. However, there are numerous missing values in water quality data due to inescapable factors like sensor failure, thereby leading to biased result for hydrological analysis and failing to support environmental governance decision accurately. A Latent Factorization of Tensors (LFT) with Stochastic Gradient Descent (SGD) proves to be an efficient imputation method. However, a standard SGD-based LFT model commonly surfers from the slow convergence that impairs its efficiency. To tackle this issue, this paper proposes a Fast Latent Factorization of Tensors (FLFT) model. It constructs an adjusted instance error into SGD via leveraging a nonlinear PID controller to incorporates the past, current and future information of prediction error for improving convergence rate. Comparing with state-of-art models in real world datasets, the results of experiment indicate that the FLFT model achieves a better convergence rate and higher accuracy.

## I. INTRODUCTION

Owing to the advancement of sensor technology, there is an increasing number of automatic water quality monitoring stations which utilize sensors to automatically detect and record various water quality metrics, including PH, dissolved oxygen and temperature [1-3]. A robust decision support for pollution abatement, water resources manager and urban planning can provided by these water quality data collected by sensors [4, 5]. However, there is a considerable number of missing values typically in water quality data due to the factors such as sensor malfunctions and network fault [6, 7]. As illustrated in Fig.1, the water quality data generated from water quality monitoring sites can be represented by a series of lists, where numerous values are unknow. Applying such water quality data with missing values can lead to biased results in statistical analysis and increase the inaccuracy and uncertainty of decision support [8]. Hence, how to accurately and fast implement imputation of missing value in water quality data becomes a thorny issue [9].

As previous research [10-16, 18-21], a latent factorization of tensors (LFT) model can accurately predict missing spatiotemporal data via modeling them into a high-dimensional and incomplete (HDI) tensor that fully preserves the structural information of observed data. Wu et al. propose spatiotemporal missing data prediction model based on LFT with instance-frequency-weighted regularization [19]. Wang et al. propose a multi-linear-algebra based concepts of tensor completion model [17]. Zhang et al. propose the time-aware model leverages the constrain of average value to predict unknow values [22]. Although such LFT models are efficient for missing data imputation, they normally suffers from slow convergence due to adopting a standard SGD as solver [24-32].

A PID controller leverages the past, current and future information to automatic control the systems [33-36]. The integral part in PID controller accumulate historic error to reduce the state error, the proportional part controller the current error and the derivative part generates a control signal according to the rate of change of the error. In resent research, it's proved that the PID module can accelerate the convergence of the SGD algorithm by refining the instance error [37-40].

Inspired by the above discovery, this paper proposes a Fast Latent Factorization of Tensors (FLFT) model. The FLFT incorporates PID module into SGD for fast and accuracy imputation of missing values in water quality data, and a special non-linear function is adopted to adjust integral and derivation parts of PID controller dynamically for obtaining high prediction accuracy. The main contributions of this paper are provided as:

a) A FLFT model. It implements fast and accurate imputation of missing value in water quality data.

b) Detailed algorithm design and analysis for FLFT. It offers detailed instructions for researchers on applying FLFT.

Empirical studies on real-world water quality data show that compared with state-of-the-art models, The FLFT achieves high convergence rate and prediction accuracy.

## II. PRELIMINARIES

### A. Water quality data tensor

Water quality data can be easily represented by an HDI tensor. The stations and their monitoring metrics can be represented by a matrix at each time slot and the HDI tensor consists of matrices in difference time slots, as shown Fig. 1.

Q. Liu is the College of Computer and Information Science, Southwest University, Chongqing, China, (email: liuqianswu@163.com).
L. Wang is the School of Artificial Intelligence Chongqing University of Education, Chongqing, China, (email: wanglan@cque.edu.cn).
B. Yang is the School of Environment and Ecology Chongqing University, Chongqing, China, (email: cq_yangbing@163.com)
H. Wu is the College of Computer and Information Science, Southwest University, Chongqing, China, (haowuf@gamil.com).

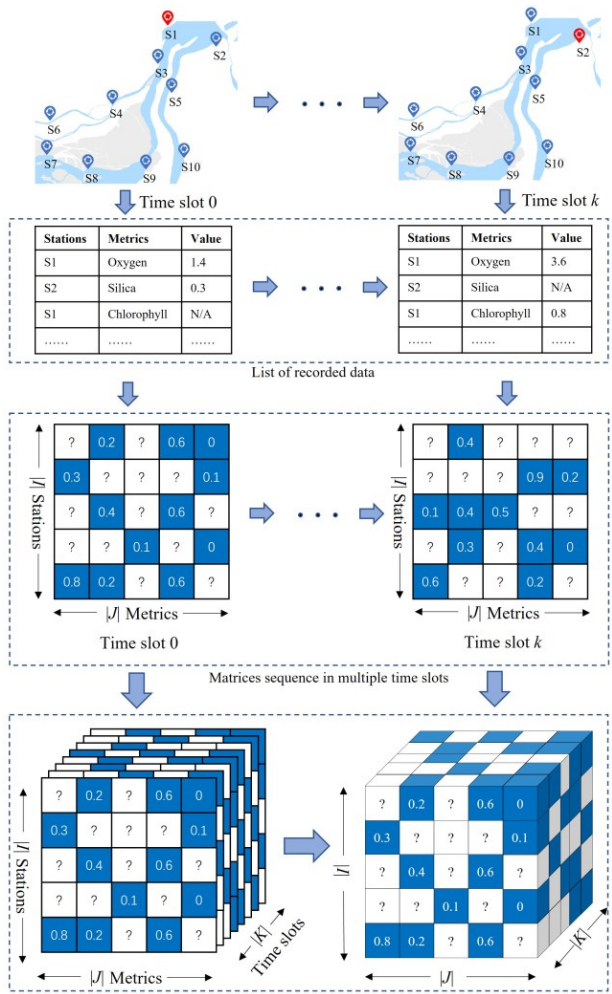

Fig. 1. The HDI tensor describing water quality data.

Definition 1 (HDI tensor): Given entity sets *I*, *J* and *K*, each of them represent the monitoring stations, the monitoring metrics of these stations and the time slots, those consist tensor. $Y^{|I| \times |J| \times |K|}$. each $y_{ijk}$ in this tensor represents the monitoring value of the $i^{th}$ station on $k^{th}$ time slot with $j^{th}$ type of metrics. Fig .2 depicted the HDI tensor.

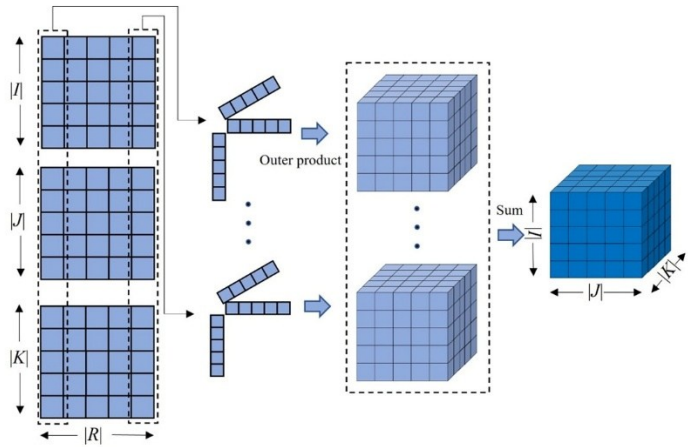

Fig. 2. Building HDI tensor by factor matrixes.

## B. Latent factorization of tensors

Definition 2 (LFT model): An LFT model builds a rank-R approximation $\hat{Y}$ to $Y$, where three latent factor (LF) matrices $S^{|I|\times|R|}$, $M^{|J|\times|R|}$, $T^{|K|\times|R|}$, are used as the low-dimensional representations of stations, metrics and time slots. Accordingly, each element in $\hat{Y}$ is formulated as:

$$\hat{y}_{ijk} = \sum_{r=1}^{R} s_{ir} m_{jr} t_{kr} \tag{1}$$

Note that the $s_{ir}$, $m_{jr}$ and $t_{kr}$ denote the single element of LF matrix S, M and T. Fig. 2 illustrate the process of building tensor.

To obtain LF matrices, on the known data set $\Lambda$ of HDI tensor, an object function measuring the difference between $y_{ijk}$ and $\hat{y}_{ijk}$ is built [23] as:

$$\varepsilon = \sum_{y_{ijk}\in\Lambda}\left(y_{ijk}-\hat{y}_{ijk}\right)^2 = \frac{1}{2}\sum_{y_{ijk}\in\Lambda}\left(y_{ijk}-\sum_{r=1}^{R}s_{ir}m_{jr}t_{kr}\right)^2 \tag{2}$$

Due to its sensitivity to initial hypotheses of S, M and T, incorporating the Tikhonov regularization into LFT is essential [41-44, 47]. Moreover, it is necessary to incorporate the linear bias into objective function to enhance the model learning ability [45-48]. Thus, by incorporating three linear bias vectors, the objective function of LFT is given

$$\begin{aligned}\hat{y}_{ijk} &= \sum_{r=1}^{R} s_{ir} m_{jr} t_{kr} + a_i + b_j + c_k \\ \varepsilon &= \frac{1}{2}\sum_{y_{ijk}\in\Lambda}\left(\left(y_{ijk}-\hat{y}_{ijk}\right)^2 + \lambda\sum_{r=1}^{R}\left(s_{ir}^2 + m_{jr}^2 + t_{kr}^2 + a_i^2 + b_j^2 + c_k^2\right)\right)\end{aligned} \tag{3}$$

Stochastic gradient descent algorithm is usually used to LFT model[49-54]. According to the principle of SGD, the learning scheme for each element of LF and linear bias can be obtained as (4).

$$\begin{cases} s_{ir}^{n+1} \leftarrow s_{ir}^n - \eta\frac{\partial f}{\partial s_{ir}^n} = s_{ir}^n + \eta\left(\left(y_{ijk}-\hat{y}_{ijk}\right)m_{jr}^n t_{kr}^n - \lambda s_{ir}^n\right) \\ m_{jr}^{n+1} \leftarrow m_{jr}^n - \eta\frac{\partial f}{\partial m_{jr}^n} = m_{jr}^n + \eta\left(\left(y_{ijk}-\hat{y}_{ijk}\right)s_{ir}^n t_{kr}^n - \lambda m_{jr}^n\right) \\ t_{kr}^{n+1} \leftarrow t_{kr}^n - \eta\frac{\partial f}{\partial t_{kr}^n} = t_{kr}^n + \eta\left(\left(y_{ijk}-\hat{y}_{ijk}\right)s_{ir}^n m_{jr}^n - \lambda t_{kr}^n\right) \\ a_i^{n+1} \leftarrow a_i^n - \eta\frac{\partial f}{\partial a_i^n} = a_i^n + \eta\left(\left(y_{ijk}-\hat{y}_{ijk}\right) - \lambda a_i^n\right) \\ b_j^{n+1} \leftarrow b_j^n - \eta\frac{\partial f}{\partial b_j^n} = b_j^n + \eta\left(\left(y_{ijk}-\hat{y}_{ijk}\right) - \lambda b_j^n\right) \\ c_k^{n+1} \leftarrow c_k^n - \eta\frac{\partial f}{\partial c_k^n} = c_k^n + \eta\left(\left(y_{ijk}-\hat{y}_{ijk}\right) - \lambda c_k^n\right) \end{cases} \tag{4}$$

The formula (4) explains the update scheme for the element s and a. The similar formula can be adopted when updating the element of M, T and the bias b and c

### III. THE FLFT MODEL

## A. A Non-linear PID controller

PID controller consists of three terms: proportional, integral and derivative. The refined error by the standard PID controller is given

$$\tilde{e}^n = K_P e_{ijk}^n + K_I \sum_{f=1}^{n} e_{ijk}^f + K_D\left(e_{ijk}^n - e_{ijk}^{n-1}\right) \tag{5}$$

$$e^n = y_{ijk} - \hat{y}_{ijk} = y_{ijk} - \sum_{r=1}^{R} s_{ir}^n m_{jr}^n t_{kr}^n - a_i^n - b_j^n - c_k^n \tag{6}$$

The refined error is formed by a combination of three terms. $K_P e_{ijk}^n$ is the proportional term and plays role of learning rate. The integral term $K_I \sum_{f=1}^{n} e_{ijk}^f$ considers the past learning information and $K_D(e_{ijk}^n - e_{ijk}^{n-1})$ is the derivative term which leverages the difference between two iterations to incorporate the trend of error.

The integral item can lead fast convergence but cause overshoot while the derivative can avoid it [55-57]. Hence, we can set a larger integral term for decreasing the steady-state error when error becomes small. Moreover, the derivative term is set small at the beginning and increases when error decreases, in order to prevent overshoot. The non-linear function formal is given

$$f(x, \alpha) = |x|^{\alpha} \cdot sign(x) \quad (0 < \alpha < 1) \tag{7}$$

The non-linear function mentioned in (7) adopts a combination of power and sign functions. The magnification of the mapping becomes higher when error gets smaller. For instance, when α equal to 0.5, the value of α will be mapping to about 0.7 while 0.1 will be mapping to about 0.3. As Fig.3 depicted, a smaller *α* means a significant amplification for small error.

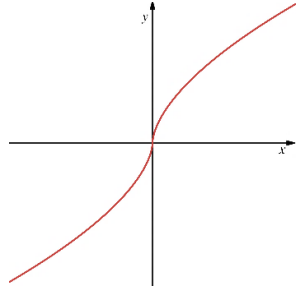

Fig. 3. The non-linear function

Now, by incorporating the non-linear function into the terms of integral and derivative in (5). the refined error ê is given

$$\tilde{e}^{(n)} = K_P e_{ijk}^{(n)} + K_I f(\sum_{i}^{n} e_{ijk}^{(i)}, \alpha_i) + K_D f(e_{ijk}^{(n)} - e_{ijk}^{(n-1)}, \alpha_d) \tag{8}$$

The $\alpha_i$ and $\alpha_d$ mentioned in formula (8) are non-linear gain parameters for integral and derivative part. Using the refined error mentioned above to replace the instant error in (4), we get the refined learning scheme for FLFT model as follows.

$$\begin{cases} s_{ir}^{n+1} \leftarrow s_{ir}^n - \eta \frac{\partial f}{\partial s_{ir}^n} = s_{ir}^n + \eta(\tilde{e}_{ijk}^n m_{jr}^n t_{kr}^n - \lambda s_{ir}^n) \\ m_{jr}^{n+1} \leftarrow m_{jr}^n - \eta \frac{\partial f}{\partial m_{jr}^n} = m_{jr}^n + \eta(\tilde{e}_{ijk}^n s_{ir}^n t_{kr}^n - \lambda m_{jr}^n) \\ t_{kr}^{n+1} \leftarrow t_{kr}^n - \eta \frac{\partial f}{\partial t_{kr}^n} = t_{kr}^n + \eta(\tilde{e}_{ijk}^n s_{ir}^n p_{jr}^n - \lambda t_{kr}^n) \\ a_i^{n+1} \leftarrow a_i^n - \eta \frac{\partial f}{\partial a_i^n} = a_i^n + \eta(\tilde{e}_{ijk}^n - \lambda a_i^n) \\ b_j^{n+1} \leftarrow b_j^n - \eta \frac{\partial f}{\partial b_j^n} = b_j^n + \eta(\tilde{e}_{ijk}^n - \lambda b_j^n) \\ c_k^{n+1} \leftarrow c_k^n - \eta \frac{\partial f}{\partial c_k^n} = c_k^n + \eta(\tilde{e}_{ijk}^n - \lambda c_k^n) \end{cases} \tag{9}$$

## IV. EXPERIMENT RESULTS AND ANALYSIS

### A. General settings

**Datasets:** There are three datasets of marine water quality monitoring data of Victoria Harbor from Hong Kong Environmental Protection Department1 which are labeled as D1-D3. D1 detect the water quality from the surface of water. It contains 129415 known entities with 24 stations and 17 metrics while the density is 0.351. D2 comes from the monitoring data in middle depth with the 121218 know entities and its density is 0.329. D3 is the data comes from the bottom of stations with 129406 know entities.

Each dataset is randomly divided with the ratio of 2:2:6 into three parts: training, validation and testing set. The parameter of rank for all model is fixed at 10 to balance the time cost and the ability of representation.

The model training process terminates for two situations that 1) the sum of the iteration counts arrived 500, or 2) the differ between two iteration is less than 1e-5 for five consecutive times.

**Evaluation Metrics:** In this experiment, we use RMSE as evaluation metrics which is commonly used in measure the representation ability of model [58-61]. The smaller RMSE means the better accuracy. $\Omega$ is the testing set from $\Lambda$.

$$RMSE = \sqrt{\frac{\sum_{y_{ijk} \in \Omega}\left(y_{ijk} - \hat{y}_{ijk}\right)^2}{|\Omega|}}$$

### B. Comparison With State-of-art Models

In this experiment, Four state-of-art models are involved to compare with FLFT model, i.e., HDOP [17]: A multi-dimensional data predict model using adaptive regularization coefficient and learning step size; WSPred [22]: A model based on CP framework leverages the average value to prevent the predicted values varying a lot; BLFT [45]: It is the stochastic gradient descent based latent factorization of tensor with bias; PLFT [56]: It is a LFT model based on SGD with a linear PID controller.

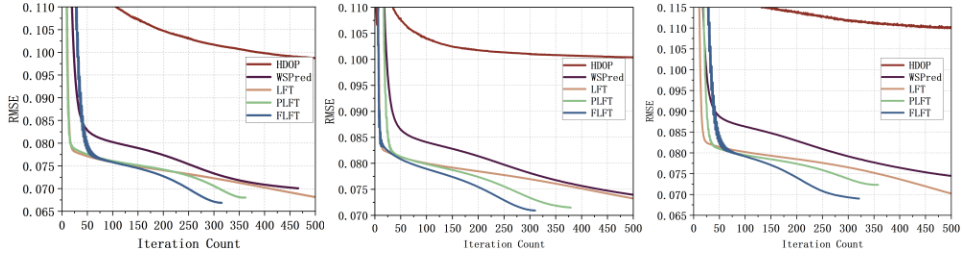

Fig. 4. Training curves of all models on D1-D3

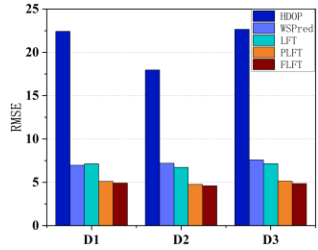

Fig. 5. The time cost of all models on D1-D3

a) FLFT model achieves better accuracy compared with its peers. As shown in Fig.4, FLFT model's RMSE is 0.066 which is lower than 0.07, 0.098, 0.068 obtained by WSPred, HDOP, LFT model, and is also lower than linear PID model's RMSE. Similarly, FLFT model has the lowest RMSE on other datasets.

b) FLFT model compared with other models gains faster convergence. In Fig.4. FLFT converges on RMSE in 321 iterations while the WSPred, PLFT converge in 461 and 360 iterations. LFT and HDOP are model consecutive training until the max iterations. FLFT also cost the minimum iterations to get convergence on other subfigures in Fig.4.

c) FLFT model compared with other models has lowest time cost. In Fig.5. it depicts the time cost of all models on D1-D3. FLFT obtains the lowest time cost on RMSE with 4.91 computational cost compared with the WSPred's 6.96, HDOP's 22.4, LFT's 7.14 and PLFT's 5.08. The time cost on D2, FLFT model only spends 4.59 computational cost while WSPred, HDOP, LFT and PLFT model use 7.19, 17.97, 6.72 and 4.74. FLFT model also obtains the lowest cost on D3.

In total, the experimental results indicate that FLFT model has the characteristic of a fast convergence rate and demonstrates highly competitive accuracy.

## V. Conclusion

To predict the missing water quality data, this paper proposes a Fast Latent Factorization of Tensors (FLFT) model. This model leverages a non-linear PID controller reconstruct prediction error, thereby improving the convergence rate. The experiment results indicate that the PLFT model obtains a higher convergence rate and great accuracy. Further, there are multiple forms of non-linear functions, we consider design new forms of non-linear function to improve convergence rate and prediction performance in future work.